\documentclass{article}

\usepackage[preprint]{nips_2018}

\usepackage[utf8]{inputenc} 
\usepackage[T1]{fontenc}    
\usepackage{hyperref}       
\usepackage{url}            
\usepackage{booktabs}       
\usepackage{amsfonts}       
\usepackage{nicefrac}       
\usepackage{microtype}      
\usepackage{graphicx}
\usepackage{float}
\usepackage{url}
\title{Overcoming catastrophic forgetting problem by weight consolidation and long-term memory}

\author{
  Shixian Wen \\
  Department of Computer Science\\
  University of Southern California\\
  Los Angeles, CA 90089\\
  \texttt{shixianw@usc.edu} \\
  \And
  Laurent Itti \\
  Department of Computer Science \\
  University of Southern California\\
  Los Angeles, CA 90089\\
  \texttt{itti@usc.edu} \\
}

\begin{document}

\maketitle

\begin{abstract}
  Sequential learning of multiple tasks in artificial neural networks using gradient descent leads to catastrophic forgetting, whereby previously learned knowledge is erased during learning of new, disjoint knowledge. Here, we propose a new approach to sequential learning which leverages the recent discovery of adversarial examples. We use adversarial subspaces from previous tasks to enable learning of new tasks with less interference. We apply our method to sequentially learning to classify digits 0, 1, 2 (task 1), 4, 5, 6, (task 2), and 7, 8, 9 (task 3) in MNIST (disjoint MNIST task). We compare and combine our Adversarial Direction (AD) method with the recently proposed Elastic Weight Consolidation (EWC) method for sequential learning. We train each task for 20 epochs, which yields good initial performance (99.24\% correct task 1 performance). After training task 2, and then task 3, both plain gradient descent (PGD) and EWC largely forget task 1 (task 1 accuracy 32.95\% for PGD and 41.02\% for EWC), while our combined approach (AD+EWC) still achieves 94.53\% correct on task 1. We obtain similar results with a much more difficult disjoint CIFAR10 task, which to our knowledge had not been attempted before (70.10\% initial task 1 performance, 67.73\% after learning tasks 2 and 3 for AD+EWC, while PGD and EWC both fall to chance level). Our results suggest that AD+EWC can provide better sequential learning performance than either PGD or EWC.
\end{abstract}

\section{Introduction}
Continual learning is a crucial step for us to design a general A.I. system, able to learn new tasks sequentially without forgetting old tasks. However, current deep learning models based on stochastic gradient descend approaches severely suffer from catastrophic forgetting [1][2], in that they often forget all old tasks after training each new one. 

Inspired by the mammalian neocortex, which relies on processes of task-specific synaptic consolidation to enable continual learning [3] [4] [5] [6], several concepts have been proposed, based on sequential Bayesian learning, which consist of applying a regularization function to a network trained by an old task to learn a new task. Many of these approaches work by finding a local minimum for task B around the local region in the parameter space that was optimized for task A, such as learning without forgetting [7], elastic weight consolidation (EWC) [8], and incremental moment matching [9]. However, restricting the parameters for task B to the local region around the optimum of task A prevents the neural network from finding other regions in remote areas of the parameter space, which might contain a better local minimum for the joint probability distribution of tasks A and B.

In our brain, the hippocampus (HPC) [10] encodes detailed information in Cornu Ammonis 3 (CA3), which does pattern separation and transforms this information into abstract high-level information, then relayed to Cornu Ammonis 1 (CA1), which does pattern completion. During weight consolidation [11] [12] [13], the HPC fuses different features from different tasks into a coherent memory trace. Over days to weeks, as memories mature, HPC progressively stores permanent abstract high-level long-term memories to remote memory storage (neocortical areas). HPC can maintain and mediate their retrieval independently when the specific memory is in need. 

Artificial neural networks are vulnerable to adversarial examples [14]. By adding a carefully computed “noise” to an input picture, without changing the neural network, one can force the network into misclassification. The noise is usually computed by backpropagating the gradient in a so-called “adversarial direction” [15]. Going to the adversarial direction, such as by using the fast gradient sign method [16], can help us generate adversarial examples that span a continuous subspace of large dimensionality (adversarial subspace). Because of “excessive linearity” in many neural networks [17][18], due to features including Rectified linear units and Maxout, the adversarial subspace often takes a large portion of the total input space. Once an adversarial input lies in the adversarial subspace, nearby inputs also tend to lie in it. 

Attack and defense researchers usually view adversarial examples as a curse of neural networks, but we view it as a gift to solve catastrophic forgetting. Points that lie inside the adversarial subspace of each class lead the neural network to misclassify input images into that class. The intersection of adversarial subspaces belonging to each class may lead neural network misclassify input images into those classes that form it. Although, we do not know how our brain represents permanent long-term memories, we propose that the intersection of the adversarial subspaces of all known classes is a representation of abstract long-term memory in our network. We want to embed the adversarial subspaces that have been found in the input space into our neural network, so that the parameters in our neural network behave like input images and span adversarial subspaces in the parameter space. Thus, we add memory units in the neural network and update their values in the adversarial direction (see methods). Each memory unit is a float tensor that can hold a float value. Each neuron contains several memory units for each task separately. Memory units play a role similar to adversarial input images, capture the essence of each class, and span an intersection of adversarial subspaces in network parameter space.  In this work, in analogy to remote memory storage and retrieval, and weight consolidation in biological brains, we design an adversarial memory net which has memory units to store and retrieve the abstract long-term memory information for each task in each neuron, and we use elastic weight consolidation (EWC) to form the joint probability representation of tasks.

Our main contribution is to view the catastrophic forgetting from a new angle. Existing approaches consist of either searching the local region around parameters that are optimal for the old tasks to find a local minimum for a new task, or storing most of the data from old tasks into a working memory and replaying it when training a new task [19]. Our approach leverages the intersection of adversarial subspaces as an abstract long-term high-level representation that captures the essence of data from old tasks. We store each intersection of adversarial subspaces from all classes of each task separately. Without any replaying of previous data, we can achieve a high accuracy on new tasks while minimally decreasing accuracy on old tasks. Our approach also is not constrained to the local region around the parameters that are optimal for the old tasks. Instead, we create a new space that is good for both old and new tasks.

We show that our adversarial memory net can be trained for disjoint MNIST and disjoint CIFAR10 tasks sequentially while minimally decreasing accuracy on old tasks. Beyond these results which use a 5-layer fully connected network, we show that our approach can also apply to CNNs, although more research is necessary to enable EWC constraints to apply to convolutional and pool layers.
    
\section{Adversarial Memory Network (AMN)}
One perhaps obvious approach to avoiding interference between sequential tasks might be to use a separate network for each task. However, such method is not scalable with the number of tasks, because there is a one-to-one mapping from network to task. Here, we add some finite number of memory units to each neuron in the fully-connected layers of the network, which remember a joint high-level representation of all classes from one task, by finding the intersection of adversarial subspaces of all classes from that task. There is no requirement for one-to-one mapping between the number of added neurons and the number of tasks.

\begin{figure}[H]
	\begin{center}
		\includegraphics[width=1\linewidth,height=4cm]{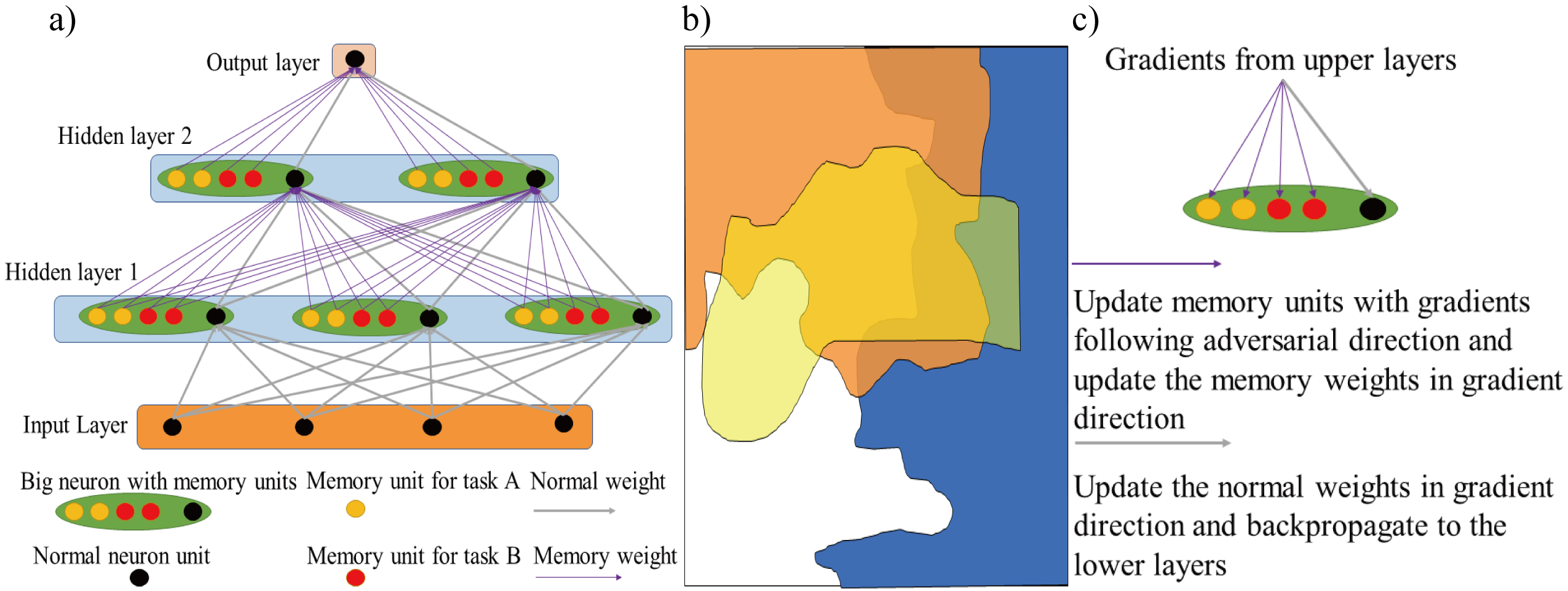}
		\caption{a) network structure of adversarial memory network. Each green big neuron contains memory units specified for each task. Yellow memory units are for task A and red memory units are for task B. The grey arrows represent normal weights corresponding to black normal neurons. The purple arrows represent memory weights corresponding to red and yellow memory units. b) Blue: adversarial subspace of Class 1. Yellow: adversarial subspace of Class 2. Orange: adversarial subspace of Class 3. Following the adversarial direction can lead us to the intersection of those 3 adversarial subspaces, and updating memory units with this information helps us to form a joint abstract high-level memory of the 3 classes. c) we update memory units with gradients following the adversarial direction, and we update the normal weights and memory weights in the gradient direction.}
	\end{center}
\end{figure}

\begin{figure}[H]
 	\begin{center}
 		\includegraphics[width=1\linewidth,height=5.5cm]{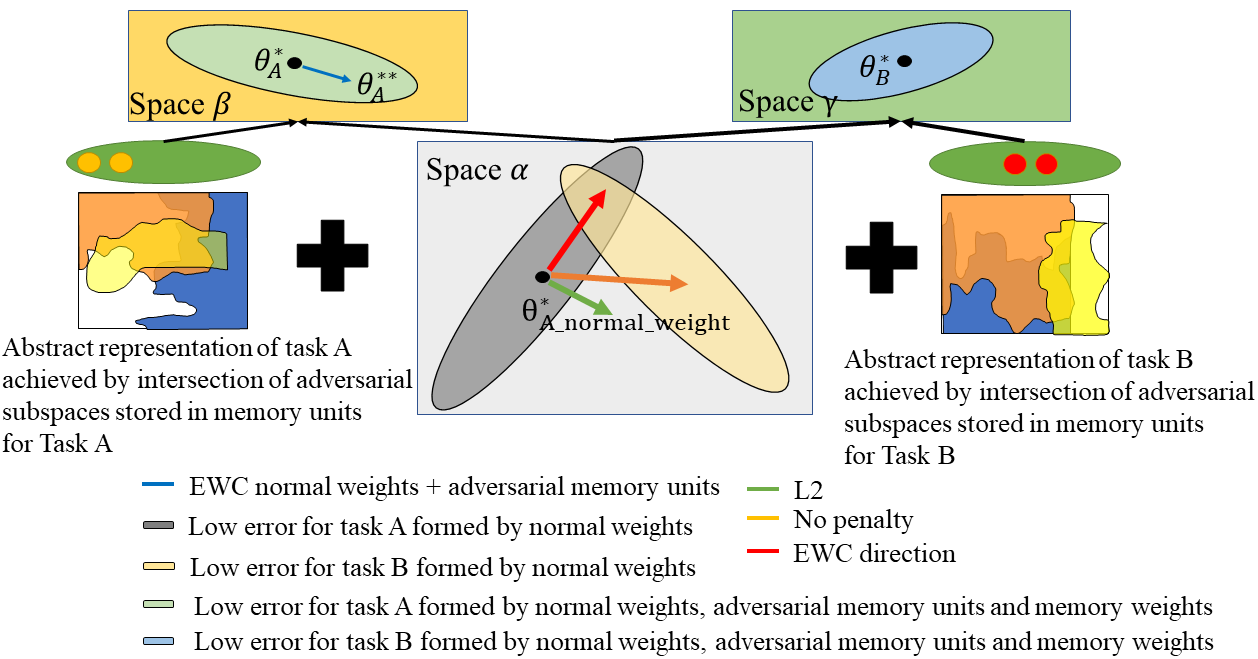}
 		\caption{High-level abstract information formed by the intersection of adversarial subspaces are stored in yellow memory units for task A and red memory units for task B separately. Space \(\beta\) is formed by the combination of normal weights, adversarial memory units (update yellow memory units with the gradient following adversarial direction for task A) and memory weights. Space \(\gamma\) is formed by the combination of normal weights, adversarial memory units (update red memory units with the gradient following adversarial direction for task B) and memory weights. Space \(\alpha\) is formed by the normal weights only and is a subspace of space \(\beta\) or space \(\gamma\). After finishing the training of task A, the parameters are \(\theta_A^*\) in space \(\beta\) and \(\theta_{A\_normal\_weight}^*\) in space \(\alpha\). \(\theta_{A\_normal\_weight}^*\) is a subset of \(\theta_A^*\). When we train task B, we restrict \(\theta_{A\_normal\_weight}^*\) following the EWC direction, we update the memory units for task B with the gradient following the adversarial direction, and we update the memory weights for task B in the gradient direction. After training task B, we get a \(\theta_A^{**}\) that is in the low error region for task A in space \(\beta\) and \(\theta_B^*\) in the low error region for task B in space \(\gamma\). \(\theta_{A\_normal\_weight}^*\) is a subset of \(\theta_B^*\).}
 	\end{center}
 \label{FIGads}
\end{figure}

In biological brains, during consolidation, abstract high-level memory is believed to be transferred from HPC to remote memory storage, such as the olfactory cortex [11]. Inspired by this idea, to have a remote memory storage, we design a big neuron that contains memory units for each task, leading to the adversarial memory network structure of figure 1.a. During training and testing, only memory units corresponding to the current task (yellow for task A or red for task B) and normal neurons (black) in the big neuron (green) will be activated. The weights for memory units are called {\em memory weights} throughout this paper. The weights for normal neurons are called {\em normal weights} throughout. What abstract high-level memory should we write into the memory units? We learned from adversarial examples that without changing a neural network, small modifications of inputs can lead to misclassification. In addition, the adversarial subspace takes a large portion of the total input space due to “excessive linearity”. Inspired by these intriguing properties of adversarial examples, we want to find the intersection of adversarial subspaces that belong to each class, so that we have a joint high-level abstract representation of different classes in each task. From figures 1.b and 1.c, we update memory units with gradients following an adversarial direction (adversarial memory units) which can help us find the intersection of the adversarial subspaces in the parameter space. Thus, we have a joint abstract high-level memory of different classes available for each task in our neural networks. One can view those memory units as storage of different small images that capture the essence of a lot of large input images. In this work, we use the Fast Gradient Sign Method \(\epsilon sign(\nabla_XJ(X,y_{target}))\), where X represents the memory units, as the adversarial direction to replace the gradient and use it to update memory units. 

\underline{Forward rules:}\\
1. Deactivate the memory units not for the current task.\\
2. \(Output =  W_{memory\_weights} * X_{memory\_units} + W_{normal\_weights} * X_{normal\_neurons} + Bias\).

\underline{Backward rules:}  $$ X_{Gradients}=\left\{
\begin{array}{rcl}
\epsilon\;sign\;(X_{Gradients})        &      & { if\;X\;is\;memory\;unit}\\
X_{Gradients}     &      & {else}\\

\end{array} \right. $$
In figure 2, high-level abstract information formed by the intersection of adversarial subspaces is stored in yellow memory units for task A and red memory units for task B separately. Space \(\beta\) is formed by the combination of normal weights, adversarial memory units (update yellow memory units with the gradient following adversarial direction for task A) and memory weights. Space \(\gamma\) is formed by the combination of normal weights, adversarial memory units (update red memory units with the gradient following adversarial direction for task B) and memory weights. Space \(\alpha\) is formed by the normal weights only and is a subspace of space \(\beta\) or space \(\gamma\). After finishing the training of task A, the parameters are \(\theta_A^*\) in the space \(\beta\) and \(\theta_{A\_normal\_weight}^*\) in space \(\alpha\). \(\theta_{A\_normal\_weight}^*\) is a subset of \(\theta_A^*\). When we train task B, 1) we restrict \(\theta_{A\_normal\_weight}^*\) by following the EWC direction, 2) we update the memory units for task B with the gradient following the adversarial direction, and 3) we update the memory weights for task B in the gradient direction. After training task B, we obtain \(\theta_A^{**}\) that is in the low error region for task A in space \(\beta\) and \(\theta_B^*\) in the low error region for task B in space \(\gamma\). \(\theta_{A\_normal\_weight}^*\) is a subset of \(\theta_B^*\). Our method thus differs from EWC because, if one only restricts the normal weights in the EWC direction in space \(\alpha\), with disjoint tasks, we cannot always find a local minimum that is good for both task A and task B in the local region of \(\theta_{A\_normal\_weight}^*\) in space \(\alpha\). Here, our new spaces \(\beta\) and \(\gamma\) allow us to obtain a good local minimum \(\theta_A^{**}\) for task A in space \(\beta\) and \(\theta_B^*\) for task B in space \(\gamma\) independently.

\section{Experiment Results}
We test our adversarial memory network with two datasets and 3 tasks for each dataset – disjoint MINST tasks and disjoint CIFAR-10 tasks (table 1). We train 3 tasks sequentially, each for 20 epochs. We show that our approach can also apply to CNN (LeNet) by replacing the fully connected layers (FC) with our adversarial memory layers, and keeping the convolution and maxpool layers. We explore 4 different training methods:

1. \underline{EWC only}: deactivate all the memory units so that our adversarial memory network becomes a fully connected network, and use elastic weight consolidation only. 

2. \underline{AD}: we use the adversarial memory units only. After finishing the training of task 1, we freeze all the normal weights, and only allow the update of memory units in adversarial direction and memory weights in gradient direction in the latter tasks.

3. \underline{EWC + AD}: normal weights are updated in EWC direction. The memory units are updated in adversarial direction and memory weights are in gradient direction.

4. \underline{PGD}: we deactivate all the memory units so that our adversarial memory network become a fully connected network, and use plain gradient descent.

We test 4 different networks (in table 2) with different hyperparameters. Note how the settings with 3 output neurons (networks 1 \& 2 in Table 2) will associate several labels with each output neuron, one per task. We use these settings for easy comparison with previous work, but we note that the settings with 9 output neurons (networks 3 \& 4) may be preferred in practice because they yield unambiguous classification results.

\begin{table}[H]
	\begin{center}
		\caption{Disjoint MINST tasks and disjoint CIFAR10 tasks. We train our Adversarial Memory Network sequentially on 3 disjoint tasks for each dataset.}
		\includegraphics[width=1\linewidth,height=2.2cm]{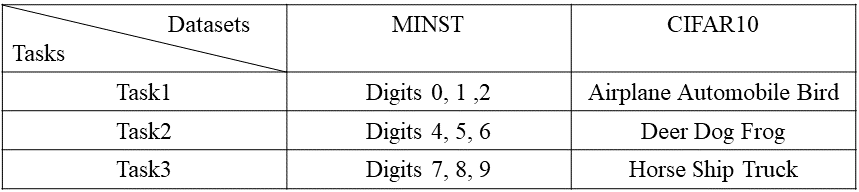}
	\end{center}
\end{table}
\begin{table}[H]
	\begin{center}
		\caption{Hyperparameters for 4 different network structures}
		\includegraphics[width=1\linewidth,height=5.8cm]{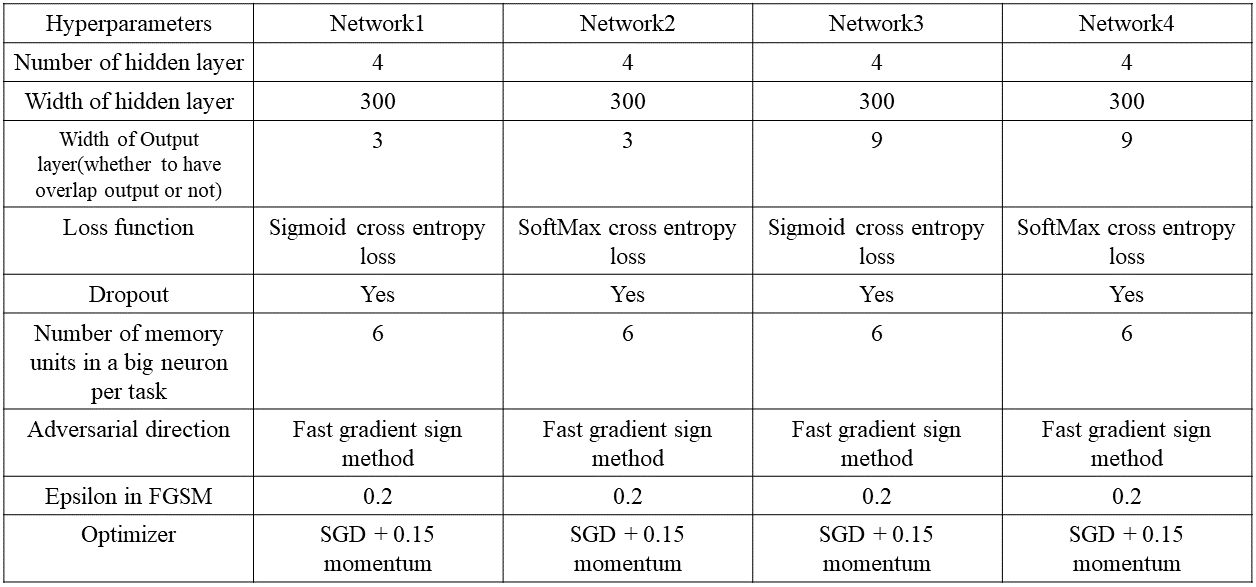}
	\end{center}
\end{table}

From figure 4 and figure 5, our adversarial memory network has adversarial memory units to form the abstract long-term memory. It uses EWC to form a joint probability distribution for sequential tasks represented by the normal memory (blue curve). It outperforms the fully connected network with EWC (red curve) or plain gradient descent (PGD; yellow curve) by a large margin in both datasets and for our 4 hyperparameter sets. In the disjoint MINST tasks, EWC + AD achieves high accuracy and barely forgets the old tasks at the same time. In the disjoint CIFAR10 tasks, we cannot achieve a high Initial accuracy because we only use a fully connected structure and do not have a convolutional structure at all. But the overall result is the same as with the MINST tasks. The sigmoid cross entropy loss converges slower than the SoftMax cross entropy loss and has a lower initial accuracy than the SoftMax cross entropy loss in disjoint CIFAR10 tasks. However, from figure 5.c, we can see that, in more complicated tasks such as disjoint CIFAR10, where tasks do not share similar low-level features, the sigmoid function may be able to pass gradients only through the active tasks since the target vector is in one-hot expression. Thus, it gives us a better accuracy for task 1 after training task 2 and task 3. The AD alone (green curve) does not forget task 1 at all, because it is frozen. But the tradeoff is that we cannot achieve high accuracies in the latter tasks. In contrast, EWC+AD can learn the latter tasks very well while only minimally decreasing task 1 accuracy. 

In figure 3.a, we demonstrate that following an adversarial direction and finding the intersection of adversarial memory subspaces is crucial, by comparing updating the memory units in the gradient direction (gradient memory units) versus the adversarial direction (FGSD). The accuracy for adversarial memory units - FGSD (blue curve) is much higher than Gradient memory units (red curve) and does not vary with the number of epochs after being trained. As a result, we argue that the intersection of adversarial subspaces in parameter space is how our neural network represents the abstract high-level information. If we store this information, we can view it as the long-term memory of our neural network.  We view the memory units as storage of the essence of a lot of input pictures from previous tasks and it is ready to retrieve the corresponding one when we test on a specific task. From figure 3.b, by varying the number of memory units in each task, we find that a low number of memory units (1 \~{} 3) in the big neuron may not be sufficient to represent high-level memory information. Yet, too many memory units (above 9) cause too much disturbance, which also decreases accuracy. Best accuracy was obtained for 6 to 9 memory units in our experiments.
\begin{figure}[H]
	\begin{center}
		\includegraphics[width=1\linewidth,height=4cm]{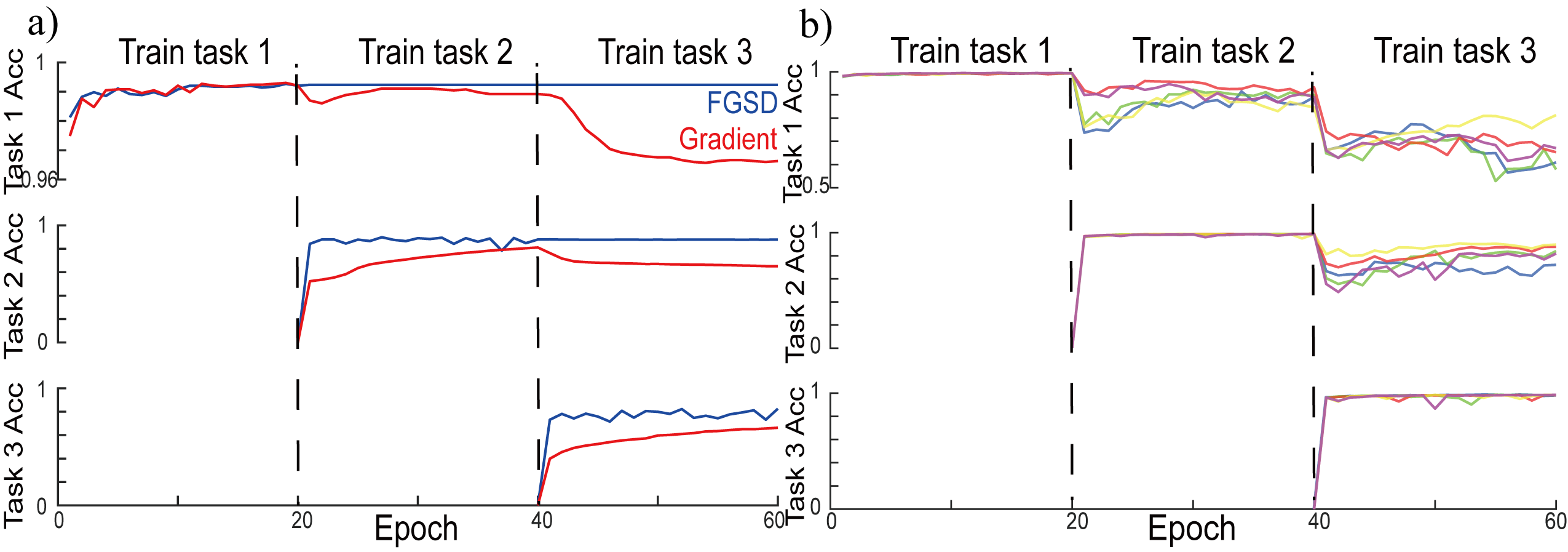}
		\caption{Disjoint MINST tasks in network 2. a) AD only. We update memory units with the gradient following the adversarial direction using the Fast Gradient Sign Method (adversarial memory units, blue curve). For comparison, we also update memory units using the normal gradient direction (gradient memory units, red curve), but this does not work as well. b) AD+EWC. Varying the number of memory units in the big neurons. Blue curve: 1 memory units. Green curve: 3 memory units. Red curve: 6 memory units. Yellow curve: 9 memory units. Magenta curve: 12 memory units.}
	\end{center}
\end{figure}

\begin{figure}[H]
	\begin{center}
		\includegraphics[width=1\linewidth,height=7.5cm]{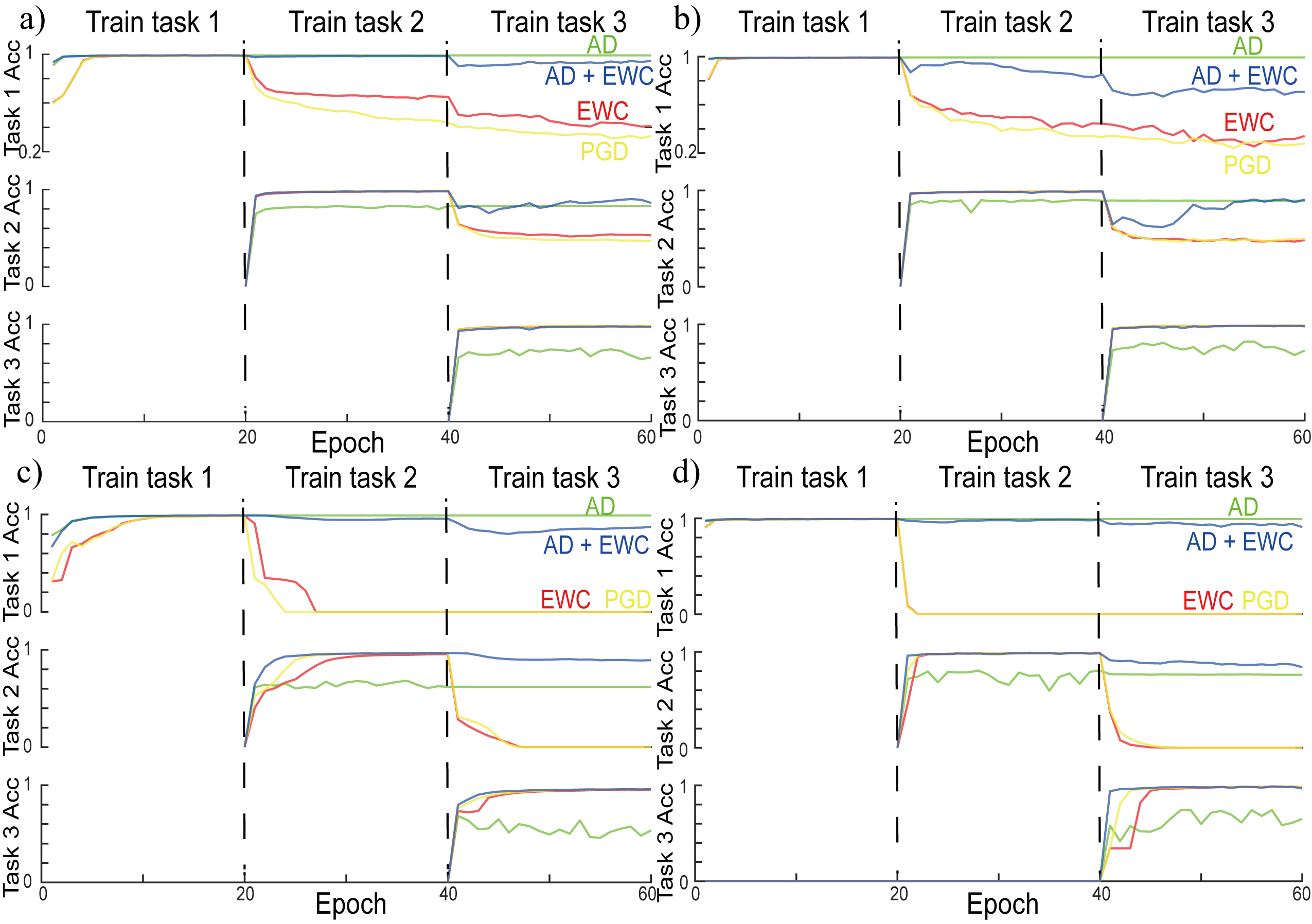}
		\caption{Disjoint MINST tasks. The red curve is using EWC alone (training method 1). The green curve is using adversarial memory units alone (training method 2). The blue curve is EWC and adversarial memory training (training method 3). The yellow curve is PGD (training method 4). Each subfigure is a) network 1, b) network 2, c) network 3, d) network 4, in Table 2.}
	\end{center}
\end{figure}

\begin{figure}[H]
	\begin{center}
		\includegraphics[width=1\linewidth,height=6.5cm]{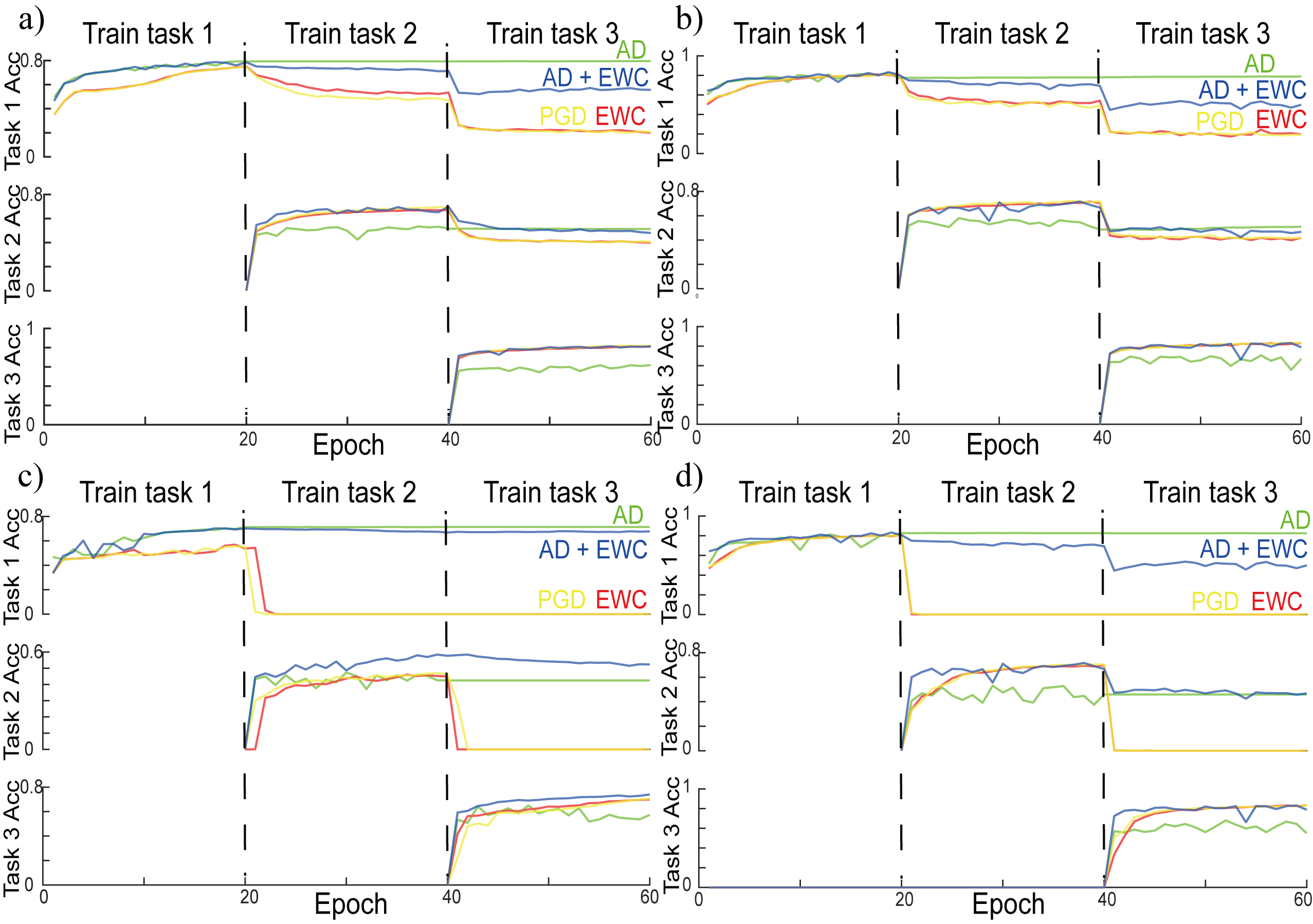}
		\caption{Disjoint CIFAR10 tasks. The red curve is using EWC alone (training method 1). The green curve is using adversarial memory units alone (training method 2). The blue curve is EWC and adversarial memory training (training method 3). The yellow curve is PGD (training method 4). Each subfigure is a) network 1, b) network 2, c) network 3, d) network 4, in Table 2}
	\end{center}
\end{figure}
\begin{figure}[H]
	\begin{center}
		\includegraphics[width=1\linewidth,height=3.5cm]{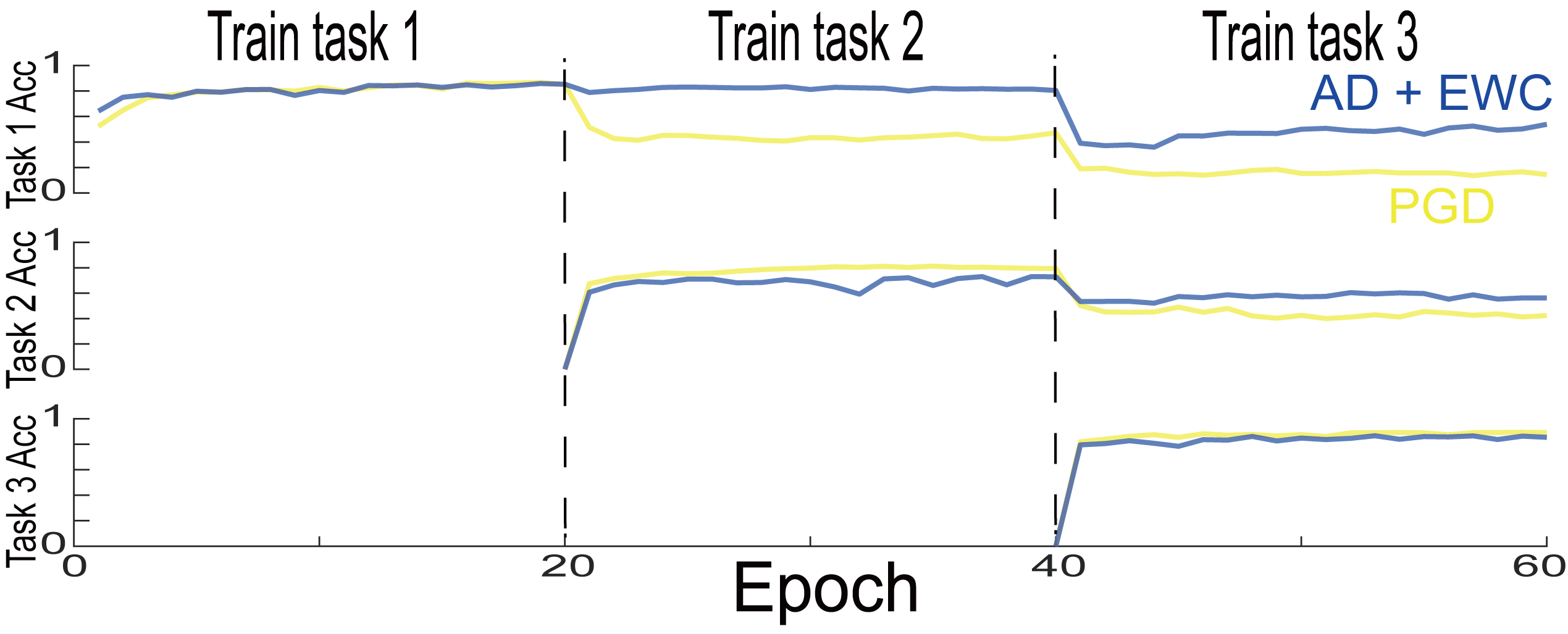}
		\caption{Disjoint CIFAR10 tasks using a convolutional network. The convolutional and pool layers are the same as LeNet, but we replace the last 3 fully connected layers by our adversarial memory layers (300 neurons for each layer). After training task 1, we freeze the parameters in the convolutional layers and use AD+EWC on adversarial memory layers (blue curve). The yellow curve is the PGD method. We use Softmax cross entropy loss with overlapped 3 outputs.}
	\end{center}
\end{figure}
Many convolutional neural networks (CNN) have some fully connected layers (FC) at the output. These are used to classify the features generated by convolutional layers. To test our method with CNNs, we use the conv and pool layers from LeNet [20] and replace its FC layers by adversarial memory layers (300 hidden units for each layer). We use Softmax cross entropy loss with overlapped 3 outputs. AD + EWC achieves high accuracy on new tasks, and achieves a better accuracy than PGD on remembering old tasks (figure 6). One difficulty with applying AD+EWC to CNNs, which will need to be addressed in future research, is to better understand how to apply EWC constraints to convolutional and pool layers. Once we know how to constrain convolutional and pool layers, we may be able to fully adapt our adversarial memory network approach to a wide range of convolutional neural network structures.

\section{Discussion}
The ability to learn sequential tasks without forgetting is a crucial step for us to design general artificial intelligence systems. We do not know how our brain stores and represents memory. Inspired by the adversarial example research and current neurobiological theories of memory consolidation, remodeling and retrieval, we present a novel network structure- adversarial memory net. It addresses the catastrophic forgetting problem in the neural network for continual learning and successfully learns the more complicated disjoint MINST tasks and disjoint CIFAR10 tasks. 
The biological inspiration is simple. Adversarial memory units in the adversarial memory net allow the network to store and retrieve the abstract high-level information represented by the intersection of adversarial subspaces of the previous tasks. This has a biological counterpart that our HPC remodels the remote memory storages and can mediate their retrieval independently. Also, by using EWC to update the normal weight, we can find a joint probability distribution of sequential tasks and has parallels with neurobiological models of synaptic consolidation in HPC. 

We can also view our approach from a mathematical point of view. Methods like EWC only apply a regularization function to a network trained by old tasks, to learn a new task based on sequential Bayesian inference, by finding a local minimum of task B around the local region of parameters space of task A. When the sequential tasks do not share similar low-level features, we usually cannot find a good joint probability distribution around the local region of parameters space of task A. Even worse is that the neural network fails to traverse other regions in the remote area of the space which might have a much better local minimum for the joint probability distribution of tasks A and B. Although we cannot revisit the data from earlier tasks in sequential learning, we can store some abstract high-level features shared by the data from earlier tasks in the memory units of our neural network and use this knowledge to classify the old test data from earlier tasks. It has been argued that the adversarial subspace takes a large portion of the total input space. Once your input is in the adversarial subspace, often all points nearby are in adversarial subspace that can fool the neural network without changing the parameters. This gives us a good representation of memory in neural network. In our adversarial memory network case, we store the adversarial gradient, which forms the intersection of adversarial subspaces in parameter space, into memory units. When we test on each task, we combine the information stored in adversarial memory units with our normal weights trained by the EWC to form new spaces (\(\beta\) and \(\gamma\) in figure 2). These new spaces usually give us a good estimate of the current test task, even though the normal weights have been modified by the EWC algorithm. This opens a new world for us that instead of trying to find the local minimal of task B in the nearby region of normal weights trained on task A, we build new curvature spaces which our experiments show have better local minima for tasks A and B, by combination of normal weights, memory weights, and adversarial memory units. If we know how to find the intersection of the adversarial subspaces of all of classes from all sequential tasks, we can use a finite number of memory units to store it and overcome the catastrophic forgetting problem completely and thoroughly. This is scalable since we do not have one to one relationship from a separate network to each task.

In our experiments, EWC alone did not prevent catastrophic forgetting, though it did perform better than PGD. Yet, our EWC implementation works well with the original permuted MINST task used by the authors of EWC. Here, we tested networks with 3 shared outputs and 9 individual outputs separately, to evaluate the influence of having different classes from different tasks share the same label. With shared outputs, as used in the original EWC work except that our dataset is not permuted MNIST, after finishing training of task 3, task 1 accuracy for both EWC and PGD fell to chance level. With individual outputs, during the training of task 2 and task 3, both EWC and PGD rapidly decreased to 0\% accuracy because the neural network cannot map task 1 to the corresponding correct outputs properly. In comparison, our EWC+AD and AD methods work well in both output scenarios.

We test sigmoid cross entropy loss and SoftMax cross entropy loss separately. When we use SoftMax cross entropy loss, both the representation and losses are mixed up, which may make it harder for the neural network to learn the sequential tasks. By using a sigmoid cross entropy loss, we may able to pass gradients only through the active tasks (figure 5.c) and obtain better performance. However, we find that without a convolutional structure, the sigmoid cross entropy loss is harder to converge compared to the SoftMax cross entropy loss. Thus, it yields a low initial accuracy for the disjoint CIFAR10 tasks. AD alone does not forget at all, because the normal weights are frozen. Because of this, AD alone cannot achieve high accuracy in the latter tasks. EWC+AD gives us a benefit that we can still achieve high accuracy in the latter tasks without decreasing the accuracy for task 1 too much. 

Since most convolutional neural networks have FC layers at the end to classify the features from the convolution layers, if we knew how to define EWC constraints for the parameters in the convolutional layers, we could replace the FC with our adversarial memory layers to adapt our method into virtually any convolutional structure.    

\subsubsection*{Acknowledgments}
This work was supported by the National Science Foundation (grant numbers CCF-1317433 and CNS-1545089), C-BRIC (one of six centers in JUMP, a Semiconductor Research Corporation (SRC) program sponsored by DARPA), and the Intel Corporation. The authors affirm that the views expressed herein are solely their own, and do not represent the views of the United States government or any agency thereof.
%

\section*{References}
\medskip
\small
[1]	French, Robert M. "Catastrophic forgetting in connectionist networks." Trends in cognitive sciences 3.4 (1999): 128-135.

[2]	McCloskey, Michael, and Neal J. Cohen. "Catastrophic interference in connectionist networks: The sequential learning problem." Psychology of learning and motivation. Vol. 24. Academic Press, 1989. 109-165.

[3]	Cichon, Joseph, and Wen-Biao Gan. "Branch-specific dendritic Ca2+ spikes cause persistent synaptic plasticity." Nature 520.7546 (2015): 180-185.

[4]	Hayashi-Takagi, Akiko, et al. "Labelling and optical erasure of synaptic memory traces in the motor cortex." Nature 525.7569 (2015): 333.

[5]	Yang, Guang, Feng Pan, and Wen-Biao Gan. "Stably maintained dendritic spines are associated with lifelong memories." Nature 462.7275 (2009): 920.

[6]	Yang, Guang, et al. "Sleep promotes branch-specific formation of dendritic spines after learning." Science 344.6188 (2014): 1173-1178.

[7]	Li, Zhizhong, and Derek Hoiem. "Learning without forgetting." IEEE Transactions on Pattern Analysis and Machine Intelligence (2017).

[8]	Kirkpatrick, James, et al. "Overcoming catastrophic forgetting in neural networks." Proceedings of the National Academy of Sciences 114.13 (2017): 3521-3526.

[9]	Lee, Sang-Woo, et al. "Overcoming catastrophic forgetting by incremental moment matching." Advances in Neural Information Processing Systems. 2017.

[10] Bakker, Arnold, et al. "Pattern separation in the human hippocampal CA3 and dentate gyrus." Science 319.5870 (2008): 1640-1642.

[11] Lesburguères, Edith, et al. "Early tagging of cortical networks is required for the formation of enduring associative memory." Science 331.6019 (2011): 924-928.

[12] Squire, Larry R., and Pablo Alvarez. "Retrograde amnesia and memory consolidation: a neurobiological perspective." Current opinion in neurobiology 5.2 (1995): 169-177.

[13	Frankland, Paul W., and Bruno Bontempi. "The organization of recent and remote memories." Nature Reviews Neuroscience6.2 (2005): 119.

[14] Szegedy, Christian, et al. "Intriguing properties of neural networks." arXiv preprint arXiv:1312.6199 (2013).

[15] Tramèr, Florian, et al. "The space of transferable adversarial examples." arXiv preprint arXiv:1704.03453 (2017).

[16] Goodfellow, Ian J., Jonathon Shlens, and Christian Szegedy. "Explaining and harnessing adversarial examples." arXiv preprint arXiv:1412.6572 (2014).

[17] Ian Goodfellow. 2016. Adversarial Examples and Adversarial Training. Stanford cs231n Lecture16 slides \url{http://cs231n.stanford.edu/slides/2017/cs231n_2017_lecture16.pdf}

[18] Tramèr, Florian, et al. "The space of transferable adversarial examples." arXiv preprint arXiv:1704.03453 (2017).

[19] Sprechmann, Pablo, et al. "Memory-based Parameter Adaptation." arXiv preprint arXiv:1802.10542 (2018).

[20] LeCun, Yann, et al. "Gradient-based learning applied to document recognition." Proceedings of the IEEE 86.11 (1998): 2278-2324.
\end{document}